\documentclass[letterpaper, 10 pt, conference]{ieeeconf}  

\IEEEoverridecommandlockouts                              

\usepackage[nolist]{acronym}
\usepackage{algorithm}
\usepackage{algorithmic}
\usepackage{amsfonts}
\usepackage{amsmath}
\usepackage{amssymb}
\usepackage{array}
\usepackage{balance}
\usepackage{bm}
\usepackage{booktabs}
\usepackage{cite}
\usepackage{eso-pic}
\usepackage[dvipsnames]{xcolor}
\usepackage{flushend}
\usepackage{glossaries}
\usepackage{graphicx}
\usepackage{graphics}
\usepackage{hyperref}
\usepackage{mathrsfs}
\usepackage{multicol}
\usepackage{multirow}
\usepackage{pgffor} 
\usepackage{pgfplots} 
\usepackage{pifont}
\usepackage{relsize}
\usepackage{scalerel}
\usepackage{siunitx}
\usepackage{stfloats}
\usepackage{textcomp}
\usepackage{tikz}
\usepackage{times}
\usepackage{upgreek}
\usepackage{url}
\usepackage{verbatim}
\usepackage{xargs}
\usepackage{xfrac}
\usepackage{xinttools}
\pgfplotsset{compat=1.18}
\usetikzlibrary{calc}
\usetikzlibrary{plotmarks}
\usetikzlibrary{svg.path}

\title{\LARGE \bf
Factorized Spatio-Temporal Convolutions\\for Human Pose Estimation from Planar Lidar}

\author{Simone Arreghini, Mirko Nava, Nicholas Carlotti, Antonio Paolillo, and Alessandro Giusti
\thanks{$^{1}$All authors are with the Dalle Molle Institute for Artificial Intelligence
(IDSIA), USI-SUPSI, Lugano, Switzerland. Corresponding author: Simone
Arreghini, {\tt\small simone.arreghini@supsi.ch}}%
\thanks{This work was supported by the European Union through the project SERMAS, by the Swiss State Secretariat for Education, Research and Innovation (SERI) under contract number 22.00247, and by the Swiss National Science Foundation, grant number 213074.}
}

\newacro{fov}[FOV]{Field of View}
\newacro{cnn}[CNN]{Convolutional Neural Network}
\newacro{fcn}[FCN]{Fully Convolutional Network}
\newacro{hri}[HRI]{Human-Robot Interaction}
\newacro{fcn}[FCN]{Fully Convolutional Network}
\newacro{tcn}[TCN]{Temporal Convolutional Network}
\newacro{nms}[NMS]{Non-maxima Suppression}
\newacro{tcn}[TCN]{Temporal Convolutional Network}

\newcommand{\paperVideo}{\href{https://youtu.be/Ay7KMmk-ESA}{\texttt{youtu.be/Ay7KMmk-ESA}}}

\newcommand{\firstpagecopyright}{%
  \AddToShipoutPictureFG*{%
    \AtPageLowerLeft{%
      \raisebox{0.32in}{%
        \hspace{0.5in}%
        \fbox{%
          \begin{minipage}{%
            \dimexpr\paperwidth-1in-2\fboxsep-2\fboxrule\relax}
            \footnotesize
            \copyright~2026 IEEE. Personal use of this material is permitted.
            Permission from IEEE must be obtained for all other uses, in any
            current or future media, including reprinting/republishing this
            material for advertising or promotional purposes, creating new
            collective works, for resale or redistribution to servers or
            lists, or reuse of any copyrighted component of this work in
            other works.
          \end{minipage}%
        }%
      }%
    }%
  }%
}

\definecolor{tiago}{HTML}{AAAAAA}%
\definecolor{model_cyan}{HTML}{87D2FF}%
\definecolor{model_purple}{HTML}{BA55D3}%
\definecolor{model_yellow}{HTML}{FFAA00}%
\definecolor{kinect}{HTML}{FF0000}%
\definecolor{lidar}{HTML}{11CC22}%
\definecolor{optitrack}{HTML}{020EBB}%
\definecolor{wong_gray}{HTML}{888888}%
\definecolor{wong_black}{HTML}{333333}%
\definecolor{wong_gold}{HTML}{E69F00}%
\definecolor{wong_cyan}{HTML}{56B4E9}%
\definecolor{wong_green}{HTML}{009E73}%
\definecolor{wong_yellow}{HTML}{F0E442}%
\definecolor{wong_blue}{HTML}{0072B2}%
\definecolor{wong_red}{HTML}{D55E00}%
\definecolor{wong_pink}{HTML}{CC79A7}%
\definecolor{wong_magenta}{HTML}{CA1963}%

\begin{document}

\firstpagecopyright
\maketitle
\thispagestyle{empty}
\pagestyle{empty}

\begin{abstract}
Localizing nearby humans and estimating their facing direction are key capabilities for safe navigation and socially aware human-robot interaction.
Many pose-estimation pipelines target cameras and 3D LiDAR or assume GPU-class compute, whereas service robots are often equipped only with omnidirectional planar LiDARs and modest onboard processors.
We address omnidirectional human detection and relative 2D pose estimation from planar LiDAR sequences with a lightweight network based on Space--Time Blocks, which explicitly separate spatial processing along scan rays from temporal aggregation across scans.
Our network processes 360$\bm{^\circ}$ LiDAR sequences to output per-ray human presence, distance, and relative orientation.
We train it via cross-modal self-supervision from a narrow RGB-D body tracker in the sensors' overlap region, removing the need for manual LiDAR labels.
Quantitative experiments show that our approach consistently outperforms a parameter-matched baseline model, reducing errors in distance (-38\%), position (-28\%), and orientation (\mbox{-15\%}).
We further benchmark on the public FROG dataset, report real-time CPU inference on a service robot, and validate with in-field demonstrations, supporting its suitability for spatial perception on computationally constrained service robots.
\end{abstract}
\section{Introduction}
\begin{figure}[!t]%
    \centering%
    \frame{\includegraphics[trim={0cm 0cm 0cm 4cm},clip=true, width=1.0\linewidth]{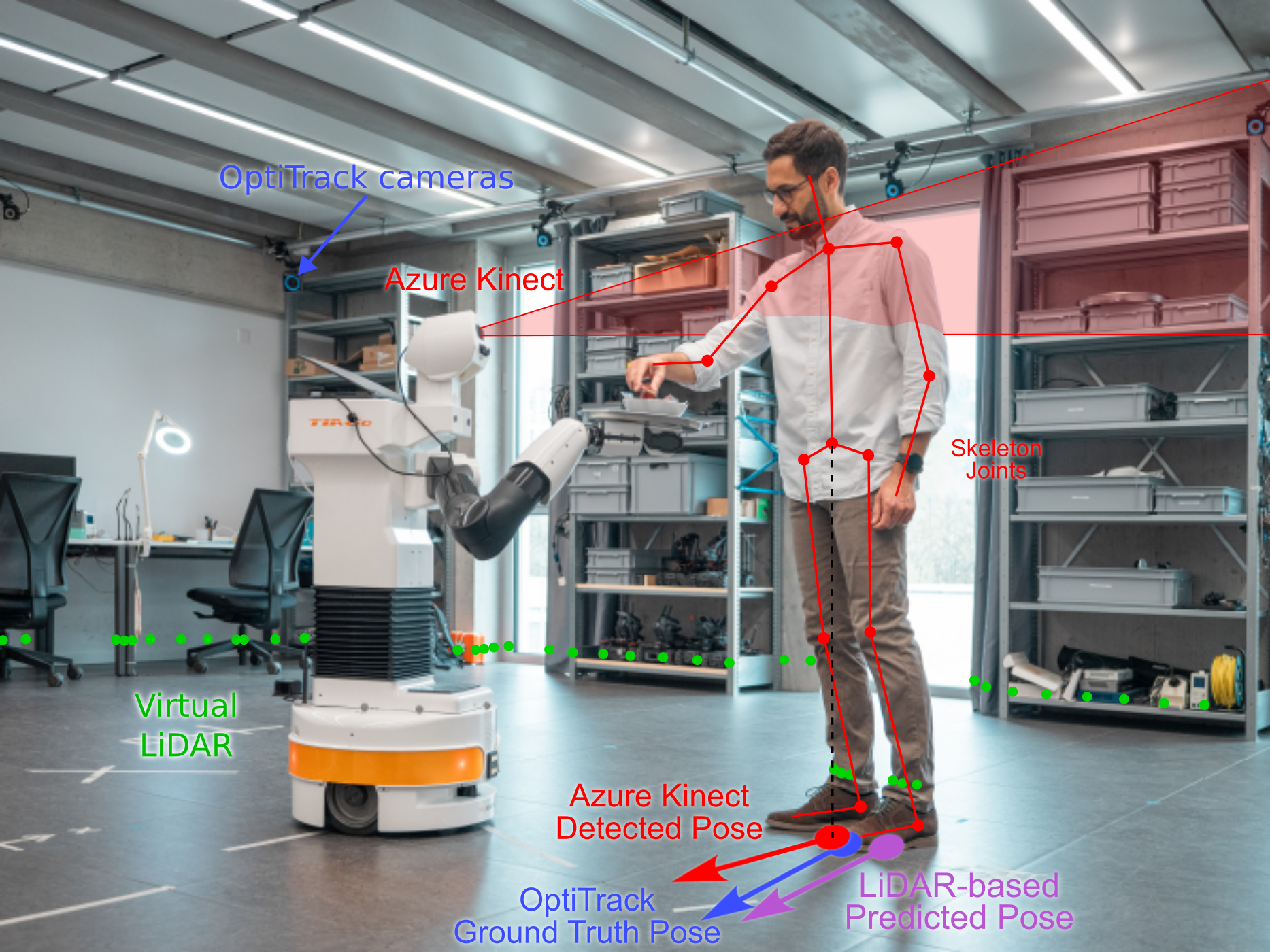}}%
    \caption{Our approach leverages the human tracking capabilities of the narrow-FOV RGB-D head camera ({\color{kinect}Azure Kinect}) as a supervision signal to train a {\color{model_purple!75!black}model} that, given planar {\color{lidar!75!black}LiDAR} scans, predicts the presence and relative 2D pose of humans around the robot. The entire pipeline relies on onboard hardware to autonomously gather and self-label data in any environment. In this laboratory, a {\color{optitrack}Motion Capture} system provides 2D ground truth poses used in the quantitative evaluation.}%
    \label{fig:experimental_setup}%
    \vspace{-16pt}
\end{figure}

The number of robots operating in spaces shared with humans is steadily increasing, ranging from large service robots~\cite{pages2016tiago} deployed in airports, hotels, and hospitals~\cite{tonkin2018design,Gonzalez:as:2021,Choi:jhmm:2020} to smaller household platforms such as robot vacuums and lawn mowers.
For these systems, perceiving nearby people and estimating their relative pose is central in enabling safe navigation and social \ac{hri} behaviors such as approaching, yielding, or initiating interaction~\cite{Arreghini:icra:2024}.
However, achieving reliable human pose perception with the sensing and computing constraints typical of mobile robots remains technically challenging.
Many such platforms rely on an omnidirectional planar (2D) LiDAR for robust, all-weather range sensing, often complemented only by a narrow-\ac{fov} camera to balance cost, power, and integration constraints~\cite{pages2016tiago,mahdi2022survey,megalingam2025cleaning}.
This setting also characterizes our robot shown in Fig.~\ref{fig:experimental_setup}, which illustrates the sensing geometry and the target quantities: human presence and relative 2D pose around the robot.
Although a planar LiDAR may provide $360^\circ$ coverage, individual scans are sparse and ambiguous compared to 3D point clouds: clutter can produce human-like returns from objects, and the approximate front-back symmetry of the human body makes facing direction difficult to infer from a single snapshot.
As a consequence, accurate estimation of position, and especially facing direction, from planar LiDAR benefits from exploiting \emph{spatio-temporal} cues across multiple scans, rather than relying on instantaneous geometry alone.
Vision-based pipelines for human detection and pose estimation are well established~\cite{paul2013human,tolgyessy2021skeleton}, and analogous approaches have been developed for 3D LiDAR point clouds~\cite{yan2020online,hayton2020cnn}.
However, transferring these solutions to the planar-LiDAR setting is nontrivial: they either rely on sensing modalities without omnidirectional coverage, or assume input representations and model capacities that do not match sparse 2D scans and typical onboard deployment constraints.

We address the task of omnidirectional human detection and relative 2D pose estimation from planar LiDAR sequences, predicting per-ray human presence, distance, and orientation.
Our design follows three pragmatic considerations.
First, many service robots carry heterogeneous sensors with complementary \ac{fov}. We exploit this by using a reliable, narrow-\ac{fov} RGB-D body-tracking output as a supervision signal to train and continuously adapt a model that operates only on wide-\ac{fov} planar LiDAR data, using a masked cross-modal loss enforced where the sensor views overlap~\cite{nava2019learning,nava2021uncertainty}.
Second, recent advances in spatio-temporal modeling show that explicitly \emph{factorizing} spatial and temporal processing can improve accuracy and optimization compared to entangling them in monolithic kernels~\cite{sun2015human,tran2018closer}.
Finally, for deployment across platforms, the perception pipeline should be sufficiently lightweight to run fully onboard, enabling continuous human awareness even on smaller robots.
While we demonstrate the system on a PAL Robotics TIAGo platform, the goal is to develop an onboard-efficient approach for omnidirectional planar-LiDAR human pose perception that can transfer easily to smaller form-factor robots with resource-constrained compute and lower-end planar LiDAR sensors~\cite{megalingam2025cleaning}.
Our \textbf{main contributions} are:
\begin{itemize}
    \item \textbf{ST-Block:} we adapt space-time factorization to circular 2D LiDAR sequences, separating circular 1D spatial convolutions from lightweight temporal convolutions that aggregate motion cues over a fixed scan history.
    \item \textbf{Masked cross-modal self-supervision:} we leverage a narrow-\ac{fov} RGB-D body tracker to gather labels and enforce a masked loss only in the area with camera-LiDAR overlap, eliminating manual annotation and enabling on-site adaptation to new environments.    
    \item \textbf{Omnidirectional planar-LiDAR pose perception on onboard compute:} we predict per-ray human presence, distance, and facing direction from planar LiDAR and validate on a motion-capture test set, with ablations, closed-loop robot trials, and a public FROG benchmark~\cite{amodeo2025frog}.
\end{itemize}
After reviewing related work in Sec.~\ref{sec:related_work}, Sec.~\ref{sec:method} presents the ST-Block model and self-supervised training, Sec.~\ref{sec:experimental_setup} describes the platform, dataset, and evaluation protocol, Sec.~\ref{sec:results} reports the results, and Sec.~\ref{sec:conclusions} concludes.
An online video illustrating the sensing setup, qualitative LiDAR pose
predictions, and closed-loop robot experiments is available at \paperVideo.
%
\section{Related Work}\label{sec:related_work}
\subsection{Sensors for Human Detection and Pose Estimation}
Vision-based human detection and pose estimation are widely used in \ac{hri}, often relying on RGB or RGB-D sensing~\cite{paul2013human,tolgyessy2021skeleton}.
In mobile robotics, LiDAR sensors are also common as they provide longer-range measurements with wide coverage and strong robustness across lighting conditions.
For people detection, both 2D and 3D LiDAR have been studied. Comparative analyses suggest that 3D LiDAR approaches are generally more resilient to occlusions, while 2D LiDAR perception retains good performance, remaining attractive due to its simpler sensing setup and favorable runtime--performance trade-offs on mobile robots with limited onboard compute~\cite{jia20222d}.
This has motivated a long line of work on people detection from 2D range scans.
Early approaches rely on hand-crafted geometric features extracted from scan segments (often leg-like patterns), combined with classical classifiers and tracking~\cite{arras2007using,chen2019pedestrian}.
More recent methods learn directly from raw scans using deep models. A common paradigm predicts person centers by combining local scan descriptors with a voting or aggregation stage, which has become a strong baseline family for 2D LiDAR detection~\cite{beyer2018deep}.
Subsequent work improves robustness by incorporating temporal context, either by aggregating features across consecutive scans or by using attention-based mechanisms to better handle motion and partial observability~\cite{jia2020dr,yang2024li2former}.
Label scarcity and domain shift are persistent challenges for methods that rely primarily on 2D LiDAR.
Self-supervised or weakly supervised pipelines mitigate these issues by exploiting cross-modal supervision~\cite{nava2019learning}, where a calibrated camera-based detector provides pseudo-labels for LiDAR data, enabling training and adaptation without manual annotation~\cite{jia2021self}.
Public benchmarks are also available, such as JRDB~\cite{martin2021jrdb} and the newer FROG dataset~\cite{amodeo2025frog}. The latter provides fully labeled knee-high 2D scans paired with robot odometry, facilitating standardized comparisons across detection approaches.
All the above methods primarily address people detection and 2D position estimation, and do not tackle the more challenging problem of estimating human facing direction.
Methods that output orientation typically rely on richer sensing or multi-modal cues, such as combining planar LiDAR with monocular images to infer oriented pedestrian pose~\cite{bu2019pedestrian}.
In contrast, our objective is an efficient pipeline for omnidirectional people detection and relative 2D pose estimation, including both position and orientation, solely from planar LiDAR at inference time.
We use a narrow-\ac{fov} RGB-D tracker as a self-supervision signal only during training and adaptation, avoiding manual annotation for each new deployment while mitigating domain shift and improving performance in the target environment.
\subsection{Space-Time Architectures}
Literature in video understanding shows that explicitly \emph{factorizing} space and time dimensions helps training and improves the performance of neural networks compared to monolithic architectures, by alternating spatial per-frame convolutions with temporal across-frame convolutions~\cite{sun2015human}.  
Early factorized designs on RGB video demonstrated that replacing a single 3D convolution with sequential 2D (spatial) and 1D (temporal) layers preserves temporal modeling while reducing parameter coupling and compute~\cite{sun2015human,tran2018closer}.
Related principles appear in sequence models for point clouds and pose estimation, where temporal convolutions and attention mechanisms are often combined to capture motion patterns over time~\cite{fan2022pstnet,lea2017temporal,liu2021graph}.
For planar LiDAR, temporal cues are particularly important because individual scans provide sparse and ambiguous information.
Our work applies the same high-level principle of separating spatial and temporal processing to omnidirectional 2D LiDAR time series.
Specifically, we introduce the Space-Time Block (ST-Block), which combines circular spatial convolutions, respecting laser scan topology, with temporal convolutions that aggregate motion patterns over time.
This design preserves the efficiency and inductive biases of convolutional models while enabling dense, per-ray predictions of human presence, distance, and relative orientation around the robot, including outside the region where training labels are directly available.
The resulting lightweight architecture aligns with the constraints of commodity service robots, where onboard compute and sensing performance are limiting factors.
%
\section{Methodology}\label{sec:method}
\begin{figure}[t]%
    \centering%
    \includegraphics[width=\columnwidth]{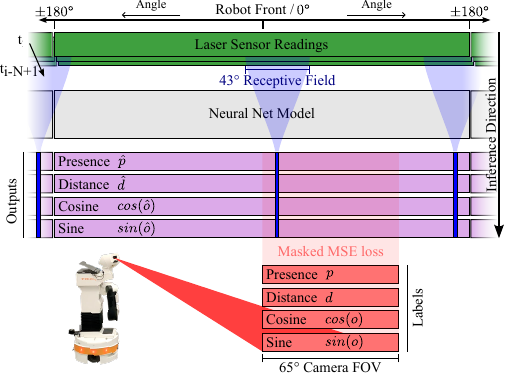}
    \caption{Our model uses a temporal window of $N$ LiDAR scans to predict the presence $\hat{p}$ of nearby people, their distance $\hat{d}$, and relative orientation $\hat{o}$ (represented by sine and cosine). Dilated circular convolutions process omnidirectional scans and yield a $43^{\circ}$ receptive field. A masked loss is enforced on predictions that overlap with the camera \ac{fov} (red shaded area).}%
    \label{fig:task_setup}%
    \vspace{-16pt}
\end{figure}
We propose a deep learning model that estimates the presence of humans around the robot, and their 2D relative pose, from a sequence of omnidirectional planar LiDAR readings. The model is trained using a supervision signal limited to the head-mounted RGB-D camera’s \ac{fov}, as shown in Fig.~\ref{fig:task_setup}. 
The model input consists of a temporal window of $N$ LiDAR scans, reprojected to the robot’s current pose using its onboard odometry. This preprocessing step alignment minimizes the impact of the robot's own motion on the input, such that static structures remain fixed across scans, while moving objects generate characteristic spatio-temporal trail-like patterns. 
Since the LiDAR provides a full $360^{\circ}$ \ac{fov}, the first and last rays are adjacent in space. This circular topology requires us to adopt circular convolutions, thereby preserving angular continuity.
For each LiDAR ray, the model predicts the likelihood of human presence $\hat{p}\in[0,1]$, the radial distance $\hat{d}$, and the person's relative orientation $\hat{o}$. We encode orientation as relative bearing, defined as the clockwise angular deviation of the facing direction with respect to the ray direction; this local (ray-centric) representation promotes generalization across the full $360^\circ$ field of view. We represent $\hat{o}$ via $\sin\hat{o}$ and $\cos\hat{o}$.
At inference time, human detections are obtained by thresholding $\hat{p}$ and applying a simple non-maximum suppression strategy.
\subsection{Space-Time Networks Architecture}
\begin{figure}[!t]%
    \centering%
    \includegraphics[width=\columnwidth]{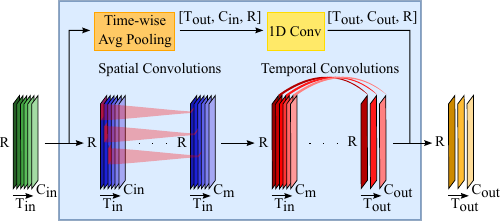}
    \caption{Space-Time Block architecture: the input (green) is processed by spatial per-ray 1D convolutions. The resulting features are then passed to the temporal layers (typically a single layer), which aggregate information along the time dimension. An optional residual path applies temporal average pooling and, if necessary, a 1D convolution to match channel dimensions. The residual output is summed with the final temporal layer output.}%
    \label{fig:st_block}%
    \vspace{-16pt}
\end{figure}

A naive approach to handle spatial and temporal information with a 1D \ac{fcn} is to stack LiDAR scans over time as additional input channels. While this strategy is simple and effective at incorporating temporal context, it does not model temporal structure, limiting the network’s ability to capture motion patterns. 
To address this limitation, we adopt the ST-Block architecture, the fundamental building unit of our network (see Fig.~\ref{fig:st_block}). Each block performs factorized spatio-temporal convolutions, separating spatial and temporal processing rather than merging them within one joint convolutional kernel.
Given an input tensor
\[
X \in \mathbb{R}^{T_{in} \times R \times C_{in}},
\]
with $T_{in}$ timesteps, $R$ rays, and $C_{in}$ channels, the block first applies 1D circular convolutions along the spatial dimension $R$ for each timestep independently:
\[
H_{t} = \text{Conv}_{\text{spatial}}(X_{t}), \quad H \in \mathbb{R}^{T_{in} \times R \times C_{m}}.
\]
The subscript $t$ indicates that the spatial convolution is applied independently at each timestep $X_{t} \in \mathbb{R}^{R \times C_{in}}$, producing an intermediate representation $H_{t}$ for that timestep. Collecting the outputs over all $T_{in}$ timesteps yields the tensor $H \in \mathbb{R}^{T_{in} \times R \times C_{m}}$.
The resulting activations $H$ are then processed by a shallow \ac{tcn} that aggregates information across timesteps while preserving causality:
\[
Z = \text{Conv}_{\text{temporal}}(H), \quad Z \in \mathbb{R}^{T_{out} \times R \times C_{out}}.
\]
Temporal convolutions use a stride equal to the kernel size $k$, which effectively compresses the sequence length by a factor of $k$, i.e., $T_{out} = T_{in}/k$.
Each ST-Block may include a residual connection: the input is downsampled in time by average pooling with factor $k$ and projected with a 1D convolution to match the channel dimension before being added to $Z$. 
Given this design, stacking ST-Blocks yields a network whose receptive field expands in both space and time. The spatial receptive field grows through dilated circular convolutions across rays, while the temporal receptive field grows through the temporal kernel size and stride used to aggregate and downsample the input history. 
After the last block,  we attach one linear layer as a prediction head, yielding the per-ray predictions.
Convolutional architectures are particularly well-suited for this setting: their translational invariance allows the detection of patterns independently of their position in the LiDAR scan. As a result, knowledge acquired in the labeled front-facing region can naturally transfer to the rest of the \ac{fov}, where direct supervision is unavailable.
To expose multi-scale information to the prediction head, we can additionally use skip connections from intermediate block outputs. Let $Z^{b} \in \mathbb{R}^{T_{out}^{b} \times R \times C_{out}^{b}}$ denote the output of block $b \in \{1,\dots,B\}$. From each block, we extract the activations at the most recent timestep, project them with a $1$D convolution $\phi^{b}$ to a common width, and concatenate across the channel dimension:
\[
S = \text{Concat}\big(\, \phi^{1}(Z^{1}_{0}), \, 
                          \phi^{2}(Z^{2}_{0}), \, 
                          \dots, 
                          \phi^{B}(Z^{B}_{0})\, \big).
\]
The full network is composed of a sequence of $B$ ST-Blocks with residual and skip connections. It takes as input a window of $N$ preprocessed planar LiDAR scans with $R$ rays each (tensor shape $R \times N$) and outputs an $R \times 4$ tensor with per-ray channels $\hat{p}$, $\hat{d}$, $\sin(\hat{o})$, and $\cos(\hat{o})$.
\subsection{Masked Loss Function and Self-supervised Training}
Labels are generated in a cross-modal pseudo-label supervision using the Azure Kinect RGB-D body tracker as an off-the-shelf component. The tracker estimates a 3D skeleton with $32$ joints per person; we retain the pelvis pose, transform it to the robot base frame, and project it onto the ground plane to obtain the 2D position and facing direction used as supervision.
This information is used to produce ray-wise labels for presence $p$, distance $d$, and relative orientation $o$ for each LiDAR ray intersecting a person. 
The presence label $p$ is $1$ if a person is detected along the ray and $0$ otherwise. Correspondingly, distance and orientation labels are assigned only to rays where a person is detected, and remain undefined otherwise.
The models operate only using planar LiDAR data and are trained using a masked regression loss, where errors are enforced only within the current camera's \ac{fov}. Likewise, distance and orientation terms are computed only on rays where people are present in the supervision signal, see Fig.~\ref{fig:task_setup}. 
Formally, the loss $\mathcal{L}$ is defined as the sum of three equally weighted terms:
\begin{equation}\label{eq:total_loss}
\mathcal{L}=\mathcal{L}_p+\mathcal{L}_d+\mathcal{L}_o,
\end{equation}
\begin{equation}
\begin{aligned}
\mathcal{L}_p &= \mathrm{MSE}(p,\hat{p}), &
\mathcal{L}_d &= \mathrm{MSE}(d,\hat{d}),\\
\end{aligned}
\end{equation}
\begin{equation}
\mathcal{L}_{\text{o}} = \tfrac{1}{2} \left[
   \text{MSE}\!\left(\sin{o}, {\sin{\hat{o}}}\right) +
   \text{MSE}\!\left(\cos{o}, {\cos{\hat{o}}}\right)
\right]
\end{equation}

Here, $p, d, o$ denote the ground-truth presence, distance, and orientation (angle), and $\hat{p}, \hat{d}, \hat{o}$ are the corresponding predictions, where $\hat{o}$ is recovered from the predicted sine and cosine components as $\hat{o} = \operatorname{atan2}(\sin{\hat{o}}, \cos{\hat{o}})$.
Since ground-truth labels are only available within the limited camera \ac{fov}, we leverage the network architecture to generalize to the entire LiDAR range, including regions with little or no direct supervision, such as behind the robot.
%
\section{Experimental Setup}\label{sec:experimental_setup}
\begin{figure}[t]
\centering
\setlength{\tabcolsep}{0.7mm}

\begin{tabular}{cc}
\frame{\includegraphics[width=.49\columnwidth]{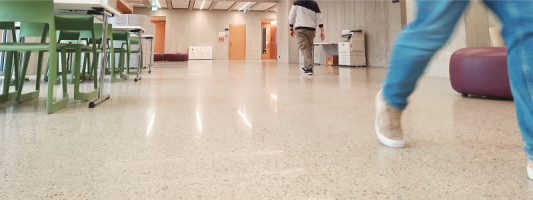}} &
\frame{\includegraphics[width=.49\columnwidth]{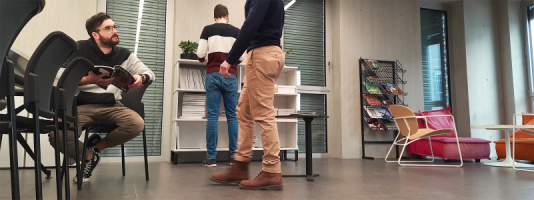}} \\
\frame{\includegraphics[width=.49\columnwidth]{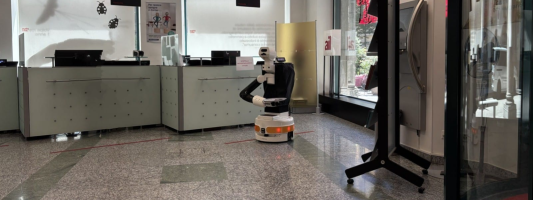}} &
\frame{\includegraphics[width=.49\columnwidth]{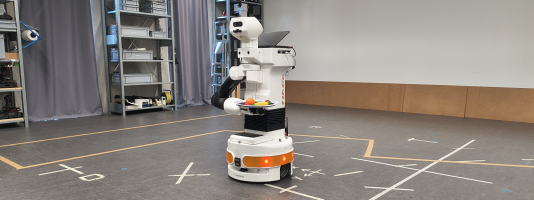}}
\end{tabular}

\caption{Dataset environments: \emph{Corridor} (top left), \emph{Break Area} (top right), \emph{Office} (bottom left), and \emph{Lab} (bottom right).}
\label{fig:scenarios}
\vspace{-16pt}
\end{figure}
We use a customized PAL Robotics TIAGo robot, featuring a differential drive base, a movable torso with a prismatic joint, a $7$~DoF manipulator, and a head capable of panning $\pm75^{\circ}$ and tilting from $-60^{\circ}$ to $45^{\circ}$ relative to the horizontal plane.
To enhance perception and coverage of its surroundings, the robot is equipped with additional sensors: 
\begin{itemize}
    \item a Microsoft Azure Kinect RGB-D camera mounted on the head, with a $65^{\circ}$ horizontal \ac{fov} and a \SI{15}{Hz} frame rate, providing reliable human tracking up to \SI{6}{m};
    \item two planar LiDARs: a front-facing Sick TiM561 integrated into the robot base at a height of \SI{95}{mm}, providing a $190^{\circ}$ \ac{fov} at \SI{15}{Hz}, and a rear-facing YDLIDAR TG15 mounted at \SI{329}{mm}, providing a $255^{\circ}$ \ac{fov} at \SI{10}{Hz}. Both sensors operate over a \SIrange{0.05}{10}{m} range and provide a distance accuracy of $\pm$~\SI{60}{mm}.
\end{itemize}
To meet the initial assumption of a single, omnidirectional, and radially symmetric sensor, we combine the two physical LiDARs into a \emph{virtual} single LiDAR.
To this end, LiDAR readings are time-synchronized at \SI{10}{Hz}, projected onto the 2D plane, and aggregated into angular bins centered at the robot's base. Each bin is assigned the distance of the closest point within it, or a default value of \SI{10}{m} if the bin contains no points.
In our setup, we chose to use $1^{\circ}$ bins, resulting in a \emph{virtual} LiDAR that provides $360$ evenly spaced rays around the robot's center. We chose this resolution to remain close to the angular resolution of lower-end planar LiDAR sensors, which supports transfer to resource-constrained platforms.
\subsection{Dataset}
We collected data across $14$ days spread over $7$ months in four environments shown in Fig.~\ref{fig:scenarios}. The first environment is \emph{Corridor}, a public transit area between university classrooms with study desks on the side and passers-by ($36$k samples); \emph{Break Area}, a large room with tables and chairs where expert individuals interact with the robot ($12$k samples); \emph{Office}, a public municipal office with booths to receive people ($6$k samples); and \emph{Lab}, a laboratory setting with expert individuals interacting with the robot ($39$k samples).
Data were collected in diverse settings to capture a wide range of human motion patterns. In the \emph{Corridor}, people—unaware of the robot—moved naturally between classrooms, sitting or standing near tables. In the \emph{Break Area} and \emph{Lab}, actors either moved around the robot or remained stationary. In the \emph{Office}, people primarily queued in front of service desks or moved while entering and exiting the space.
During data collection, the robot's base motion and the head panning were randomized to increase data variability and the area covered by the camera. The robot followed random trajectories while avoiding collisions with the exception of the \emph{Corridor} and \emph{Office} environments, where the robot base was manually controlled for safety reasons.
In all environments, we recorded body joints from the Azure Kinect and scans from the two LiDARs. Additionally, a portion of the \emph{Lab} environment ($8$k samples) provides ground truth poses for people and the robot at \SI{100}{Hz} from an OptiTrack motion capture system featuring $18$ cameras.
Data were recorded onboard the robot in the form of ROS2 bag files containing streams of timestamped messages. We use message timestamps to synchronize all modalities to the slowest topic (the rear LiDAR running at \SI{10}{Hz}), providing multi-sensor readings per timestep. Each sample includes the two LiDAR scans, the RGB-D tracker output, the robot's internal state and odometry, and, when available, motion-capture measurements. Samples are stored in chronological order within each recording, thus preserving temporal consistency.
The data split was performed by separating the recordings from completely different days, rather than randomly, to ensure that the evaluation reflects realistic generalization. 
The training set includes all samples from \emph{Corridor} ($36$k), $6$k samples from \emph{Break Area}, $4$k samples from \emph{Office}, and $28$k samples from \emph{Lab}, totaling $73.5$k samples. The validation set comprises the remaining $6$k samples from \emph{Break Area}, $2$k samples from \emph{Office}, and $3$k samples from \emph{Lab}, for a total of $11.5$k samples. The test set contains the $8$k samples from the \emph{Lab} environment recorded with additional OptiTrack data.
This collection extends the dataset introduced in Sixth-Sense~\cite{arreghini2026sixth}; its public release is available at \href{https://doi.org/10.5281/zenodo.14936069}{\texttt{10.5281/zenodo.14936069}}.
\subsection{Evaluation Setup}
All models operate on the same input data and are evaluated under identical conditions. 
LiDAR observations are aggregated over a temporal window of \SI{3}{s}, corresponding to $30$ consecutive scans at \SI{10}{Hz}, and preprocessed as described in Sec.~\ref{sec:method}. Each scan of the virtual sensor provides $360$ rays covering the full $360^{\circ}$ around the robot, yielding an input tensor of size $360 \times 30$.
For each ray, the models predict human presence, distance, and orientation. Furthermore, to ensure a fair comparison, both the baseline and the proposed ST-Block architectures are configured with the same effective receptive fields: $43$ rays (degrees) in space and $30$ timesteps (\SI{3}{s}) in time. 
\paragraph*{Baseline}  
As baseline, we use the 1D fully convolutional architecture introduced in Sixth-Sense~\cite{arreghini2026sixth}. The model processes the LiDAR history by stacking scans from different timesteps as additional channels, so convolutional kernels slide along the spatial dimension while reading all temporal information indiscriminately.
Each convolution has $32$ channels with LayerNorm and GELU activation, using circular padding to respect the angular topology. 
The first four layers use a kernel size of $3$, followed by one layer with a kernel size $5$ and two layers with a kernel size $7$. All convolutions use a stride $1$, so the receptive field grows only through kernel size and dilation, which is set to $1$ in the first three layers and to $2$ in the remaining four. All layers use $32$ feature channels until the final linear layer, which are mapped to the $4$ output channels.
We designed this baseline to isolate the effect of the proposed ST-Block factorization on this task, while keeping the receptive field comparable.
\paragraph*{ST-Block model}  
Our proposed architecture achieves the same receptive fields through a factorized design: within each ST-Block, we first apply \emph{spatial} convolutions along the ray dimension (as in the baseline) and then apply a \emph{temporal} convolution that aggregates information across timesteps.
Our implementation employs three ST-Blocks. To match the baseline spatial receptive field, we reuse the same number and configuration of spatial convolutional layers, but apply them only along the spatial (ray) dimension rather than mixing space and time. These spatial layers are distributed as follows: the first block contains three spatial convolutional layers, while the second and third blocks contain two layers each.
After each block's spatial convolutions, a temporal convolution aggregates and downsamples features along the time axis. We use progressively increasing temporal kernels with sizes $2$, $3$, and $5$.
\paragraph*{Model Training}
All networks are trained using the Adam optimizer with a learning rate of $1 \times 10^{-4}$ and a batch size of $64$ on the masked loss defined in Eq.~\eqref{eq:total_loss} for $250$ epochs. For each model replica, we select the best parameters as those achieving the lowest validation loss.  
To improve generalization, input data is augmented by adding Gaussian noise and by mirroring both the LiDAR readings and the corresponding labels across the robot's sagittal plane.  
\subsection{Public Benchmark: FROG}
We benchmark our planar-LiDAR people detector on the public FROG dataset~\cite{amodeo2025frog}, using the official evaluation protocol.
FROG provides knee-high planar LiDAR scans with ground-truth 2D person positions in the robot front-facing $180^{\circ}$ field of view, together with robot odometry, which we use in the same motion-compensation preprocessing described in Sec.~\ref{sec:method}.
Each scan contains $720$ rays over $180^{\circ}$ (4 rays per degree), in contrast to our dataset, which uses $180$ rays over the same $180^{\circ}$ area (1 ray per degree). 
Since FROG does not provide orientation annotations, we restrict this benchmark to presence $\hat{p}$ and distance $\hat{d}$ prediction.
%
\section{Results}\label{sec:results}
Metrics are computed on the $8$k samples from the \emph{Lab} environment with ground-truth human poses available from the motion capture system, and averaged over three replicas of each model configuration to cancel out the effect of random seed initialization.
For each output ray, the pose of a person is obtained by applying thresholding and non-maximum suppression, and by projecting the ray onto the 2D ground plane.

We report the following metrics: $\text{AP}$ is the area under the precision-recall curve, also called average precision. 
To analyze models at comparable operating points, we consider metrics obtained by thresholding each model with a value yielding a recall of $80\%$: precision $\text{P}_{\text{80}}$, distance error $\text{E}_{\text{d}}$, relative distance error $\text{E}_{\text{d}}^{\text{rel}}$, position error $\text{E}_{\text{p}}$, orientation error $\text{E}_{\uptheta}$, and the Pearson correlation coefficient $\uprho_{\uptheta}$ between predicted and ground-truth orientation.
\begin{table*}[thbp]
\centering
\setlength{\tabcolsep}{4pt} 
\renewcommand{\arraystretch}{1.2} 
\caption{Model performance on the test set, metrics averaged over three replicas per row.}
\begin{tabular}{lccrrrrrrr}
\toprule
\multirow{2}{*}{Architecture}
& \multirow{2}{*}{\shortstack{ST-Block \\ Residuals}}
& \multirow{2}{*}{\shortstack{Skip \\ Connections}}
& \multirow{2}{*}{$\text{AP} \uparrow$}
& \multicolumn{6}{c}{Metrics obtained with a threshold leading to 80\% recall} \\
\cline{5-10}
& & &
& \multicolumn{1}{c}{$\text{P}_{\text{80}}$ [\%] $\uparrow$}
& \multicolumn{1}{c}{$\text{E}_{\text{d}}$ [cm] $\downarrow$}
& \multicolumn{1}{c}{$\text{E}_{\text{d}}^{\text{rel}}$ [\%] $\downarrow$}
& \multicolumn{1}{c}{$\text{E}_{\text{p}}$ [cm] $\downarrow$}
& \multicolumn{1}{c}{$\text{E}_{\uptheta}$ [deg] $\downarrow$}
& \multicolumn{1}{c}{$\uprho_{\uptheta}$ [\%] $\uparrow$} \\
\midrule
Random Pred         & $-$ & $-$ & 0.1  & 0.6  & 21.5 & 8.1 & 33.0 & 90.9 & -0.2 \\
Baseline            & $-$ & $-$ & 48.5 & 73.8 & 10.7 & 4.8 & 13.2 & 37.5 & 84.4 \\
Ours (\verb|--|)   & \ding{55} & \ding{55} & \textbf{54.8} & 76.5 &  7.6 & 3.4 & 10.6 & 33.1 & 86.5 \\
Ours (\verb|R-|)   & \ding{51} & \ding{55} & 51.3 & 77.6 &  8.7 & 3.9 & 11.5 & 32.4 & 86.9 \\
Ours (\verb|-S|)   & \ding{55} & \ding{51} & 52.6 & \textbf{79.4} &  \textbf{6.6} & \textbf{2.9} &  \textbf{9.5} & 33.3 & 86.5 \\
Ours (\verb|RS|)   & \ding{51} & \ding{51} & 51.4 & 78.1 &  7.1 & 3.2 &  9.9 & \textbf{32.0} & \textbf{87.2} \\
\bottomrule
\end{tabular}
\label{tab:ablation_study}
\vspace{-12pt}
\end{table*}
\subsection{ST-Block Outperforms The Baseline}
We compare our approach against the fully convolutional network baseline, which mixes spatial and temporal information implicitly. 
As shown in Table~\ref{tab:ablation_study}, the simplest variant of our model, Ours (\verb|--|), outperforms the baseline across all metrics:
the distance error drops from $\text{E}_{\text{d}}=10.7$\, cm to $7.55$\, cm ($-29.4\%$), and position error from $\text{E}_{\text{p}}=13.2$\, cm to $10.6$\, cm ($-20\%$), while there are smaller improvements in precision in human detection from $\text{P}_{\text{80}}=73.8\%$ to $76.5\%$, orientation error $\text{E}_{\uptheta}$ from $37.5^{\circ}$ to $33.1^{\circ}$ ($-11.7\%$), and circular pearson correlation $\uprho_{\uptheta}$ from $84.4\%$ to $86.5\%$. 
These improvements come with a modest increase in model size, from $32{,}356$ parameters for the baseline to $40{,}100$ for our variants ($+24\%$), demonstrating a better performance at the same scale when compared to the baseline.
The first row of Fig.~\ref{fig:lab_performance} shows a successful detection case in the test set with three people simultaneously present, one in front of and two on the side, all correctly detected.
\subsection{ST-Block Ablation Study}
\begin{figure}[!t]%
    \centering%
    \includegraphics[width=0.32\linewidth]{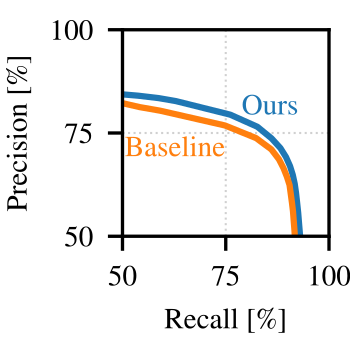}%
    \hfill
    \includegraphics[width=0.32\linewidth]{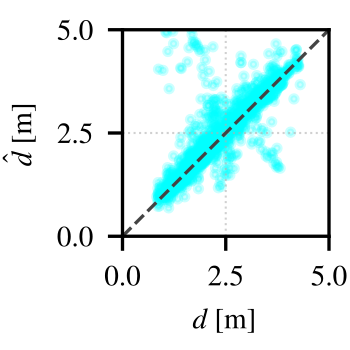}
    \hfill
    \includegraphics[width=0.32\linewidth]{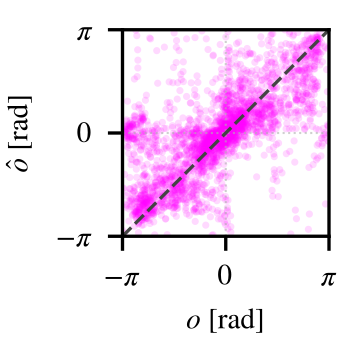}%
    \caption{PR curve comparing our ST-Block model \texttt{(-S)} with the baseline (left). Our model achieves a consistently better precision--recall trade-off, indicating more effective use of the temporal history for people detection. Scatter plots of Ours \texttt{(-S)} against ground truth for distance (center) and orientation (right) on the test set: predictions of human distance exhibit high correlation with ground truth, while orientation predictions remain more challenging, reflecting the inherent ambiguity of estimating human facing direction from planar LiDAR data.}
    \label{fig:our_model_performances}%
    \vspace{-16pt}
\end{figure}

We analyze the model performance, reported in Tab.~\ref{tab:ablation_study}, under different configurations: with or without residual connections inside each ST-Block (\texttt{R}), and with or without skip connections from block outputs to the prediction head (\texttt{S}).
The configuration with skip connections and no residuals, Ours \texttt{(-S)}, yields the lowest position errors with $\text{E}_{\text{p}}=9.5$\,cm and $\text{E}_{\text{d}}=6.6$\,cm. 
By contrast, adding residual connections inside the ST-Blocks improves orientation estimation at the cost of position estimation, reducing $\text{E}_{\uptheta}$ to $32.0^{\circ}$, and increasing $\uprho_{\uptheta}$ to $87.2\%$.
For qualitative results, we adopt Ours \texttt{(-S)} as our reference model, whose performance is shown in Fig.~\ref{fig:our_model_performances}.
\subsection{Benchmark on FROG dataset}
\begin{table}[thbp]
\centering
\setlength{\tabcolsep}{4pt} 
\renewcommand{\arraystretch}{1.2} 
\caption{Detection performance and runtime on FROG dataset.}
\label{tab:frog_performances}
\begin{tabular}{lcccrc}
\toprule
\multirow{2}{*}{Method}
& \multirow{2}{*}{$\text{AP}$}
& \multicolumn{1}{c}{Peak $F_1$} 
& \multicolumn{1}{c}{EER} 
& \multicolumn{1}{c}{Runtime} 
& \multirow{2}{*}{System} \\
& 
& \multicolumn{1}{c}{[\%] $\uparrow$}
& \multicolumn{1}{c}{[\%] $\downarrow$}
& \multicolumn{1}{c}{[ms] $\downarrow$}
& \\
\midrule
LFE-Peaks~\cite{amodeo2025frog}      & 65.6 & 70.7 & 70.7 & 1.76  & GPU \\
LFE-PPN~\cite{amodeo2025frog}        & 69.2 & 69.5 & 69.5 & 1.49  & GPU \\
Ours (Low Resolution)                & 65.9 & 70.9 & 70.9 & 2.87  & \textbf{CPU} \\
Ours (High Resolution)               & 66.4 & 72.0 & 72.0 & 3.36  & \textbf{CPU} \\
DROW3 ($T{=}1$)~\cite{beyer2018deep} & 73.9 & 72.0 & 71.8 & 13.08 & GPU \\
DR-SPAAM ($T{=}1$)~\cite{jia2020dr}  & 73.7 & 72.1 & 71.8 & 13.95 & GPU \\
DR-SPAAM ($T{=}5$)~\cite{jia2020dr}  & 75.6 & 73.6 & 73.4 & 13.99 & GPU \\
\bottomrule
\end{tabular}
\par\vspace{0.5em}\noindent
\footnotesize
\parbox{\linewidth}{%
AP = Average Precision; Peak $F_1$ = maximum $F_1$ over the decision-threshold sweep; EER = Equal Error Rate; Runtime = end-to-end inference time (preprocessing, inference, postprocessing); System: Our runtimes are measured on the CPU onboard TIAGo (Intel Core i7-10700, 16\,GB RAM); FROG runtimes are from~\cite{amodeo2025frog} (Intel Core i9-9900X, 128\,GB RAM, NVIDIA TITAN RTX 24\,GB).
}
\normalsize
\vspace{-16pt}
\end{table}
To study the impact of angular resolution, we train two variants of our \verb|RS| model:
(i) \emph{Ours (Low Resolution)} downsamples each scan to $180$ rays with no architectural changes; and
(ii) \emph{Ours (High Resolution)} preserves the FROG original $720$-ray input by prepending an adapter layer (32 channels, kernel size $5$, stride $4$, no dilation) that maps the input to the $180$-ray internal representation expected by our network.
In this benchmark, we use the \verb|RS| variant as a conservative choice for runtime, since it is the most computationally demanding configuration among our ST-Block ablations while remaining representative of the full model.
We evaluate using the official FROG benchmarking suite with association distance $d=\SI{0.5}{m}$, and report average precision (AP), Peak $F_1$, equal error rate (EER), and end-to-end runtime (preprocessing, inference, postprocessing).
Table~\ref{tab:frog_performances} reports detection performances and runtime on the FROG~\cite{amodeo2025frog} dataset.
In terms of accuracy, \emph{Ours (Low Resolution)} is competitive with the lightweight baselines provided with FROG (LFE-Peaks and LFE-PPN), while the high-resolution adapter yields only a limited improvement in AP and Peak $F_1$.
This indicates that, for coarse human awareness in common mobile service robot scenarios, increasing angular resolution beyond 1 ray/degree offers diminishing returns relative to its additional computation.
DROW3 and DR-SPAAM achieve higher AP under the FROG benchmark, but the corresponding runtimes are only reported on a powerful GPU system. 
From a deployment perspective, both of our variants run in real time on a robot CPU without any tuning or pipeline optimization, making the approach suitable for CPU-only service-robot deployment.
\subsection{Self-supervised adaptation}\label{sec:selfsup_adaptation}
To evaluate the importance of adapting to new environments and robot platforms, we perform online self-supervised adaptation starting from the low-resolution \verb|RS| variant on FROG, where the model predicts presence, distance, and orientation (since FROG does not provide orientation labels, we set both orientation targets to zero).
We then evaluate this pretrained model on our test set under the same operating conditions as Tab.~\ref{tab:ablation_study}, selecting the decision threshold that yields $80\%$ recall.
As expected for methods relying on 2D LiDAR scans, which provide limited semantic content and often yield ambiguous observations, direct transfer is poor (Tab.~\ref{tab:frog_selfsup_improvement}).
Next, we simulate deployment on a new robot by partitioning our dataset into three sequential chunks, each representing one day of operation, and use them to self-supervise the model, producing checkpoints after Day~1, Day~2, and Day~3.
As shown in Tab.~\ref{tab:frog_selfsup_improvement}, self-supervised adaptation consistently improves $\text{AP}$ and $\text{P}_{80}$, reduces position error, and learns orientation estimation from scratch, rapidly adapting the model to the operating conditions and reaching the best performance reported in Tab.~\ref{tab:ablation_study}.
\begin{table}[t]
\centering
\setlength{\tabcolsep}{4pt} 
\renewcommand{\arraystretch}{1.2} 
\caption{Impact of self-supervised adaptation starting from a model pretrained on FROG, metrics evaluated on our dataset.}
\label{tab:frog_selfsup_improvement}
\begin{tabular}{lcccc}
\toprule
\multirow{2}{*}{Time}
& \multirow{2}{*}{$\text{AP} \uparrow$}
& \multicolumn{1}{c}{$\text{P}_{\text{80}}$}
& \multicolumn{1}{c}{$\text{E}_{\text{p}}$}
& \multicolumn{1}{c}{$\text{E}_{\uptheta}$} \\
& 
& \multicolumn{1}{c}{[\%] $\uparrow$}
& \multicolumn{1}{c}{[cm] $\downarrow$}
& \multicolumn{1}{c}{[deg] $\downarrow$} \\
\midrule
Pretrain
& $12.4 \pm 4.5$
& $13.5 \pm 6.1$
& $28.7 \pm 1.3$
& $92.4 \pm 0.4$ \\
\midrule
Day 1
& $24.5 \pm 2.5$
& $51.6 \pm 5.6$
& $18.1 \pm 0.6$
& $48.6 \pm 0.7$ \\
Day 2
& $40.2 \pm 1.0$
& $65.1 \pm 1.0$
& $11.4 \pm 0.3$
& $37.3 \pm 1.3$ \\
Day 3
& $\mathbf{52.6 \pm 1.8}$
& $\mathbf{79.4 \pm 1.5}$
& $\mathbf{9.5 \pm 0.2}$
& $\mathbf{33.3 \pm 1.1}$ \\
\bottomrule
\end{tabular}
\par\vspace{0.5em}\noindent
\footnotesize
\parbox{\linewidth}{%
Metrics are computed by placing the model in the same conditions used in Tab.~\ref{tab:ablation_study}. Values are averaged over $3$ independent runs for each step.}
\normalsize
\vspace{-16pt}
\end{table}%
\subsection{In-field Demonstration}
\begin{figure*}[thbp]
    \centering
    \frame{\includegraphics[trim={10cm 11.45cm 0cm 9.5cm},clip=true,height=0.239\textwidth]{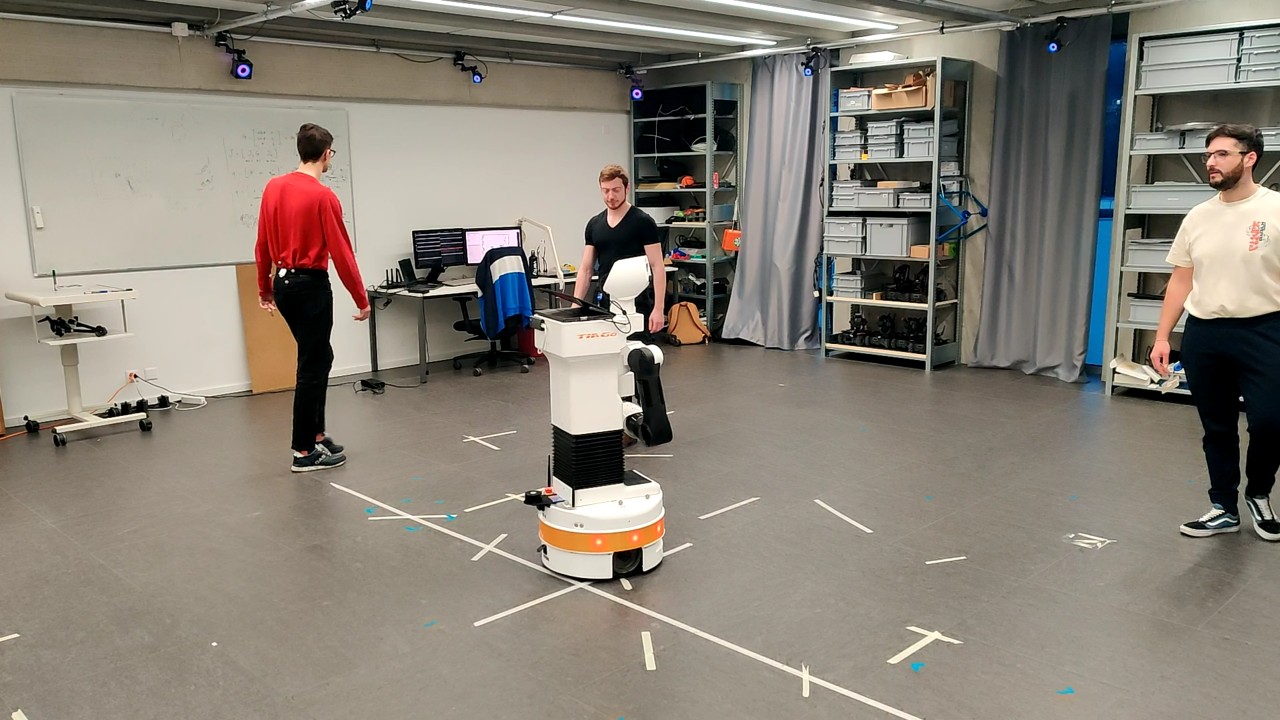}}%
    \hfill
    \frame{\includegraphics[trim={2.45cm 7.58cm 2.4cm 6.35cm},clip=true,height=0.239\textwidth]{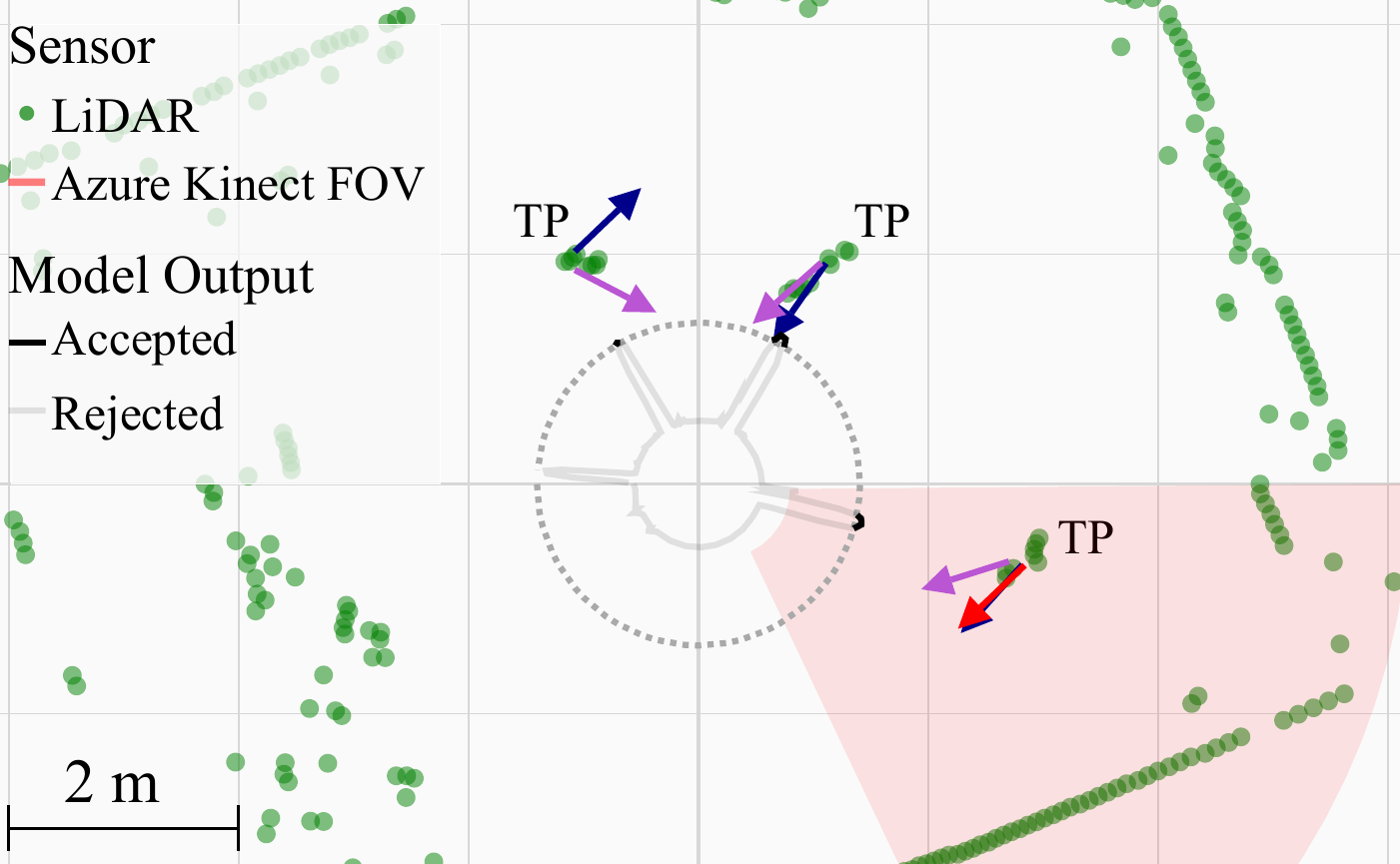}}%
    \\[1.5mm]
    \frame{\includegraphics[trim={0cm 6.0cm 0cm 2.5cm},clip=true,height=0.239\textwidth]{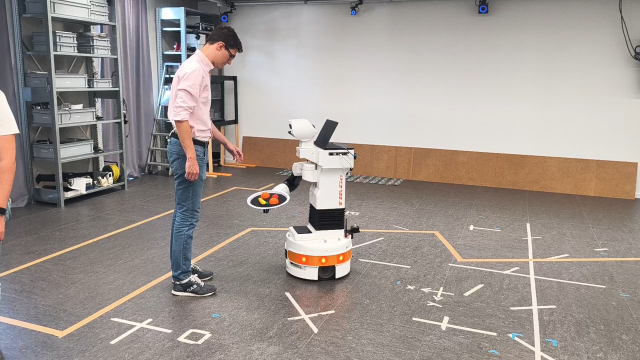}}
    \hfill
    \frame{\includegraphics[trim={2.0cm 7.4cm 2.0cm 6.0cm},clip=true,height=0.239\textwidth]{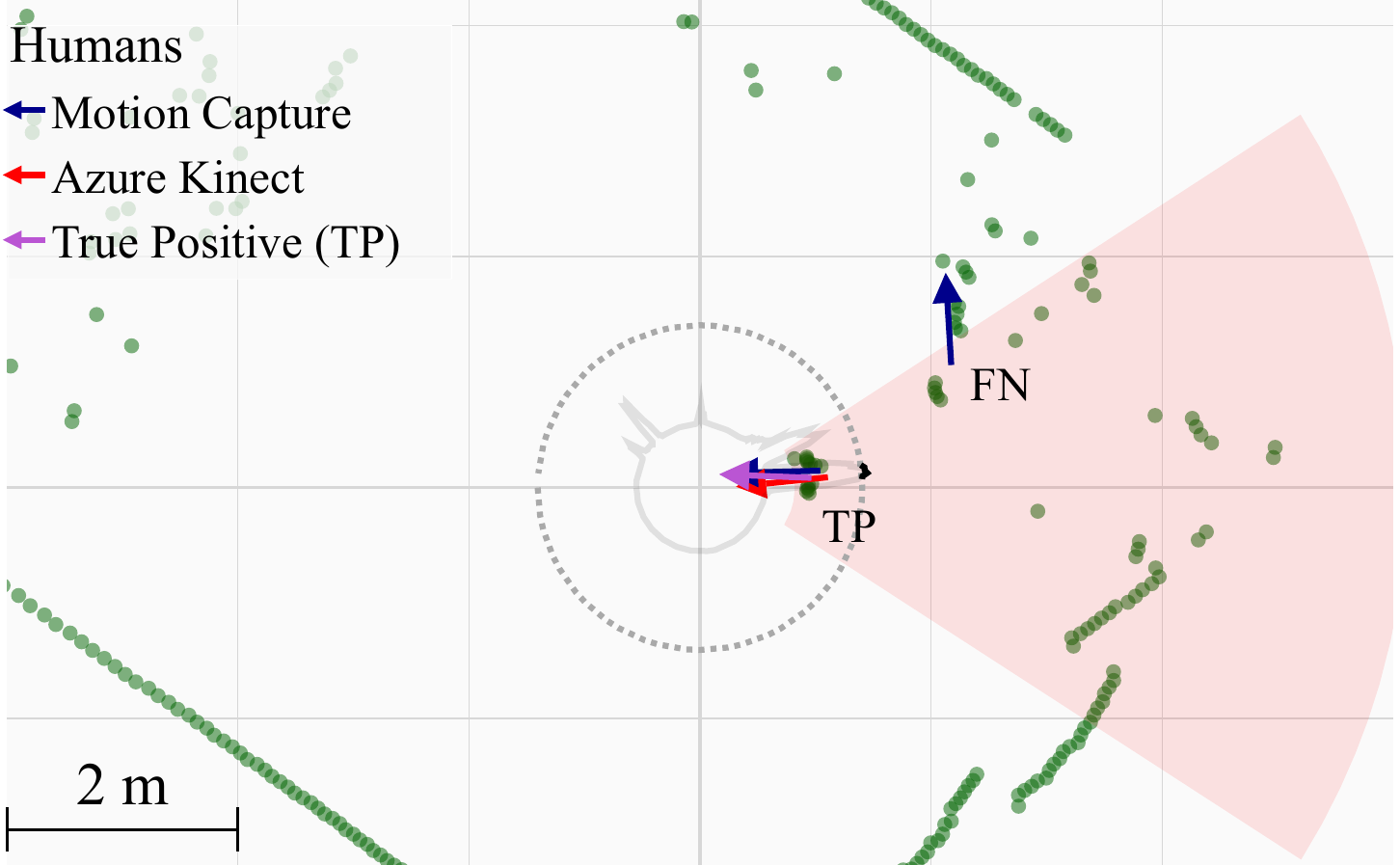}}
    \caption{%
    Third-person view of the environment (left) and corresponding top view (right) depicting the {\color{lidar!75!black}LiDAR} scan; {\color{kinect}camera \ac{fov}} and detected pose arrows; {\color{optitrack}Motion Capture} ground truth pose arrows. The predicted presence $\hat{p}$ is shown as a {\color{wong_gray!75!black}gray line} when below the same detection threshold chosen on validation to target $80\%$ recall (dashed circle centered on the robot), or black otherwise. Purple arrows represent model predictions for {\color{model_purple!75!black}true positives (TP)}.}
    \label{fig:lab_performance}%
    \vspace{-16pt}
\end{figure*}

Thanks to its omnidirectional input, our model provides continuous estimates of nearby human presence and relative pose around the robot. We use a closed-loop in-field demonstration to illustrate how this perception module can support a simple robot behavior: a service robot detects nearby people, approaches them, and performs an offering gesture similar to a waiter, receptionist, or deliveryman.
Specifically, our model performs people detection in the robot's surroundings, on top of which we apply a Kalman filter to track individuals over time and mitigate noisy predictions. 
Among the tracked people, the robot selects as the target the closest person facing the robot if closer than \SI{2.5}{m}.
The robot first rotates to face the target, then drives toward them and, once at \SI{1.5}{m} distance, extends the arm to execute the offering gesture, as shown in Fig.~\ref{fig:lab_performance} (bottom).

The figure illustrates a challenging case: the model correctly detects the closer person, while a second person behind them is partially occluded in the LiDAR scan and is not detected, since multiple people cannot be predicted on the same ray.
Overall, the experiment illustrates that the translational invariance of the fully convolutional design enables the model to generalize beyond the supervised camera field of view, producing stable detections also in regions where direct supervision was not available during training.
\section{Conclusions}\label{sec:conclusions}
We presented a lightweight, fully convolutional \emph{Space--Time Block} network for omnidirectional human detection and relative 2D pose estimation from planar LiDAR sequences.
The proposed model is an architectural instantiation of factorized spatio-temporal processing for 2D range-scan time series: it combines circular spatial convolutions, which respect the scan's angular topology, with temporal convolutions that aggregate motion cues over a fixed history window.
To avoid manual annotation and enable on-site deployment, we train the LiDAR model using masked cross-modal self-supervision from the output of a narrow-\ac{fov} RGB-D body tracker (3D skeleton/joint estimates), enforcing losses only where the camera and LiDAR fields of view overlap.
In quantitative experiments with motion-capture ground truth, ST-Block variants consistently outperform a parameter-matched 1D fully convolutional baseline that mixes space and time implicitly: at an operating point targeting \SI{80}{\percent} recall, the best configuration improves detection precision and reduces distance, position, and orientation errors (up to \SI{38}{\percent}, \SI{28}{\percent}, and \SI{15}{\percent} reductions, respectively).

A benchmark on the public FROG dataset further indicates that the approach remains competitive for 2D people detection while running in real time on a CPU, making it ideal for service robots equipped with planar LiDARs.
Limitations include reduced performance in heavy occlusions and in multi-person situations where multiple individuals fall on the same rays, as well as the inherent ambiguity of estimating facing direction from sparse 2D geometry.
In future work, we plan to (i) extend the approach from per-frame detection to multi-target people tracking and (ii) incorporate richer social cues, such as interaction intent.



\bibliographystyle{IEEEtran}
\bibliography{references}  

\end{document}